\documentclass{article}
\PassOptionsToPackage{numbers, compress}{natbib}

\usepackage[final]{new_in_ML}

\usepackage[utf8]{inputenc} 
\usepackage[T1]{fontenc}    
\usepackage{hyperref}       
\usepackage{url}            
\usepackage{booktabs}       
\usepackage{amsfonts}       
\usepackage{nicefrac}       
\usepackage{microtype}      
\usepackage{xcolor}         
\usepackage{graphicx}
\usepackage{subfig}
\usepackage{multirow}
\usepackage{balance}

\title{NutritionVerse-Synth: An Open Access Synthetically Generated 2D Food Scene Dataset for Dietary Intake Estimation}

\author{
Saeejith Nair$^{1}$ \quad Chi-en Amy Tai$^{1}$ \quad Yuhao Chen$^{1}$ \quad Alexander Wong$^{1,2}$\\
$^1$University of Waterloo, Waterloo, Ontario, Canada\\
$^2$Waterloo Artificial Intelligence Institute, Waterloo, Ontario, Canada\\
{\tt\small {\{smnair, amy.tai, yuhao.chen1, a28wong\}}@uwaterloo.ca}
}

\begin{document}

\maketitle

\begin{abstract}
Manually tracking nutritional intake via food diaries is error-prone and burdensome. Automated computer vision techniques show promise for dietary monitoring but require large and diverse food image datasets. To address this need, we introduce NutritionVerse-Synth (NV-Synth), a large-scale synthetic food image dataset. NV-Synth contains 84,984 photorealistic meal images rendered from 7,082 dynamically composed 3D scenes. Each scene is captured from 12 viewpoints and includes perfect ground truth annotations such as RGB, depth, semantic, instance, and amodal segmentation masks, bounding boxes, and detailed nutritional information per food item. We demonstrate the diversity of NV-Synth across foods, compositions, viewpoints, and lighting. As the largest open-source synthetic food dataset, NV-Synth highlights the value of physics-based simulations for enabling scalable and controllable generation of diverse photorealistic meal images to overcome data limitations and drive advancements in automated dietary assessment using computer vision. In addition to the dataset, the source code for our data generation framework is also made publicly available~\footnote{Project website:  \url{https://saeejithnair.github.io/nvsynth}}.
\end{abstract}

\section{Introduction}
\vspace{-0.1in}
Diet-related diseases are a major global health concern~\cite{malnutrition-qol}, necessitating accurate assessment of nutritional intake to guide interventions and policies. However, conventional manual approaches like food diaries are burdensome and prone to bias~\cite{kipnis2003structure}. Emerging computer vision techniques show promise for automated dietary monitoring, but progress remains constrained by the lack of suitable datasets. Prior real-world food datasets are limited in diversity of views or lack high-quality annotations~\cite{10.1145/3347448.3357172,thames2021nutrition5k}. While synthetic data can help overcome these limitations through controlled generation, no large-scale, photorealistic synthetic food dataset currently exists.

\begin{figure}[h]
    \begin{center}
        \includegraphics[width=\linewidth]{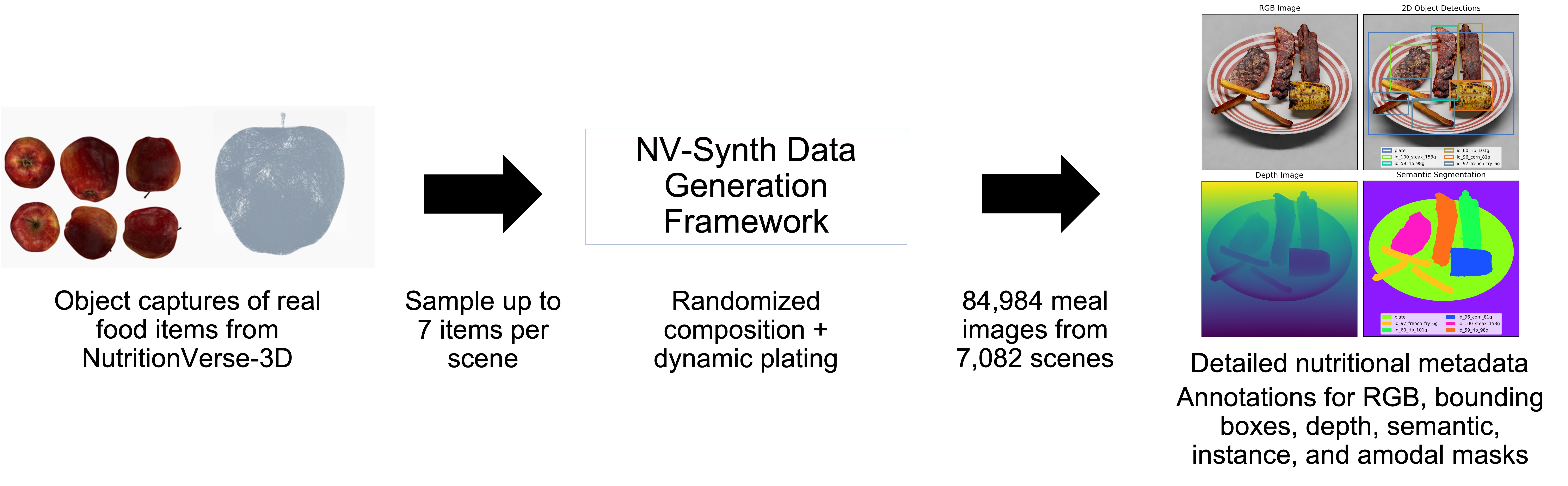}
    \end{center}
    \caption{Overview of pipeline used to generate NV-Synth dataset. We leverage 3D object capture assets of real food items from NutritionVerse-3D~\cite{nutritionverse-3d} and generate pixelwise-perfect ground truth annotations of randomly composed, dynamically plated 3D food scenes that are synthetically generated and automatically annotated using Nvidia Omniverse's Isaac Sim~\cite{issac-sim} simulator.}
    \label{fig:nvsynth_pipeline}
\end{figure}

\begin{figure}[h]
    \begin{center}
        \includegraphics[width=0.8\linewidth]{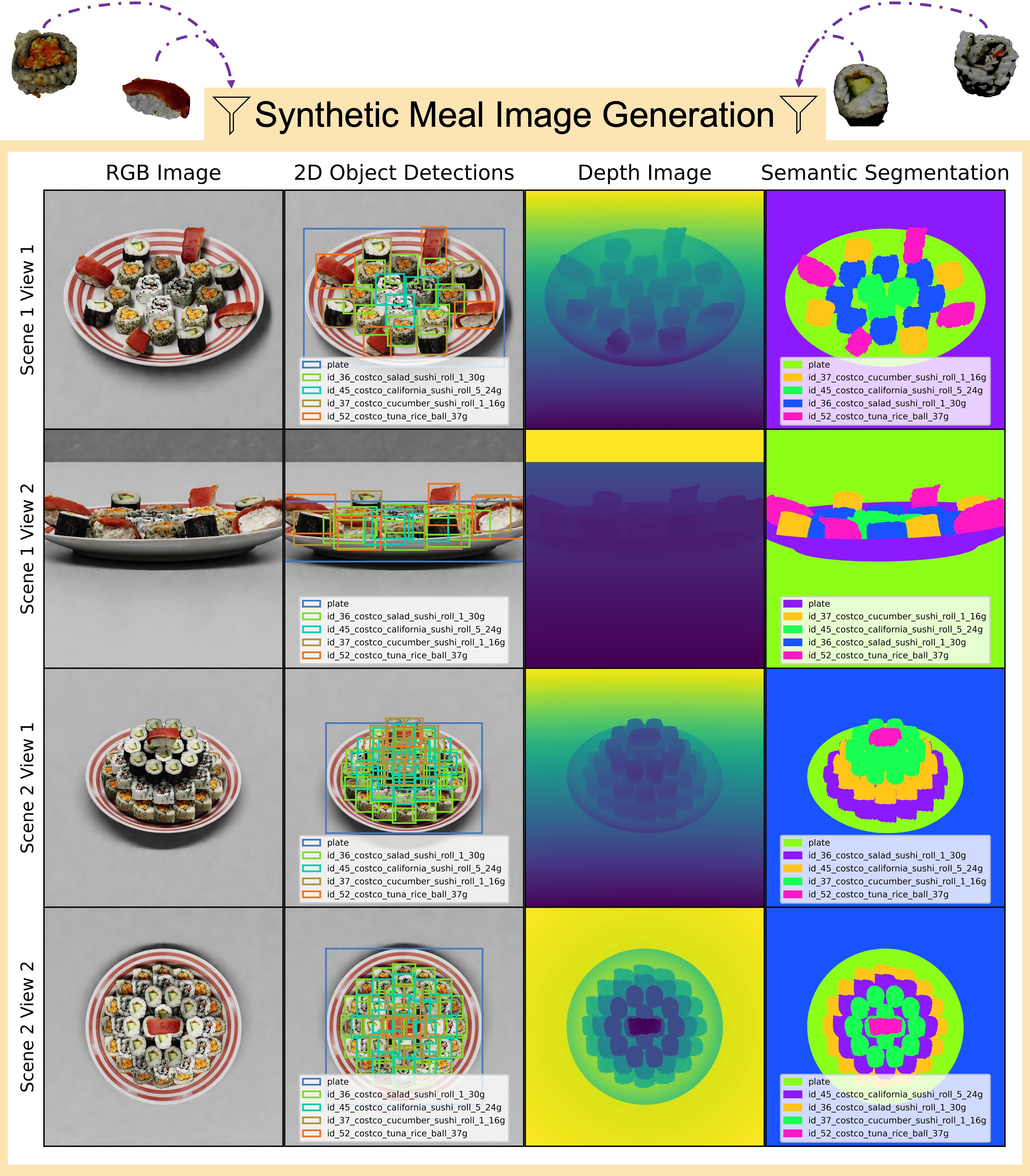}
    \end{center}
    \caption{Examples of procedurally plated 3D scenes that can be quickly generated with our framework using a human-in-the-loop. Each scene is shown from 2 viewpoints.}
    \label{fig:procedural_plating_multi_modal}
\end{figure}

To address this gap, we introduce NutritionVerse-Synth (NV-Synth), comprising 84,984 photorealistic 2D food images algorithmically rendered from 7,082 procedurally generated 3D food scenes. As shown in Figure~\ref{fig:nvsynth_pipeline}, our pipeline leverages high-quality 3D food assets from NutritionVerse-3D\cite{nutritionverse-3d} and the Isaac Sim physics engine to create realistic synthetic scenes and generate high-resolution multi-modal ground truth annotations like RGB images, depth images, object-level bounding boxes, semantic, instance, and amodal segmentation masks, as well as corresponding nutritional metadata. NV-Synth contains 105 food types rendered from 12 viewpoints per scene, enabling dietary assessment of simulated scenarios that can closely mimic how users might capture images of their meals in the real world. As the largest public synthetic food dataset, NV-Synth provides a critical resource for dietary monitoring research. Furthermore, the configurability of our framework also allows systematically expanding the scope and richness of the dataset over time.

\section{Methodology}
\begin{figure}[h]
    \centering
    \begin{minipage}{.5\textwidth}
        \centering
        \includegraphics[width=.9\linewidth]{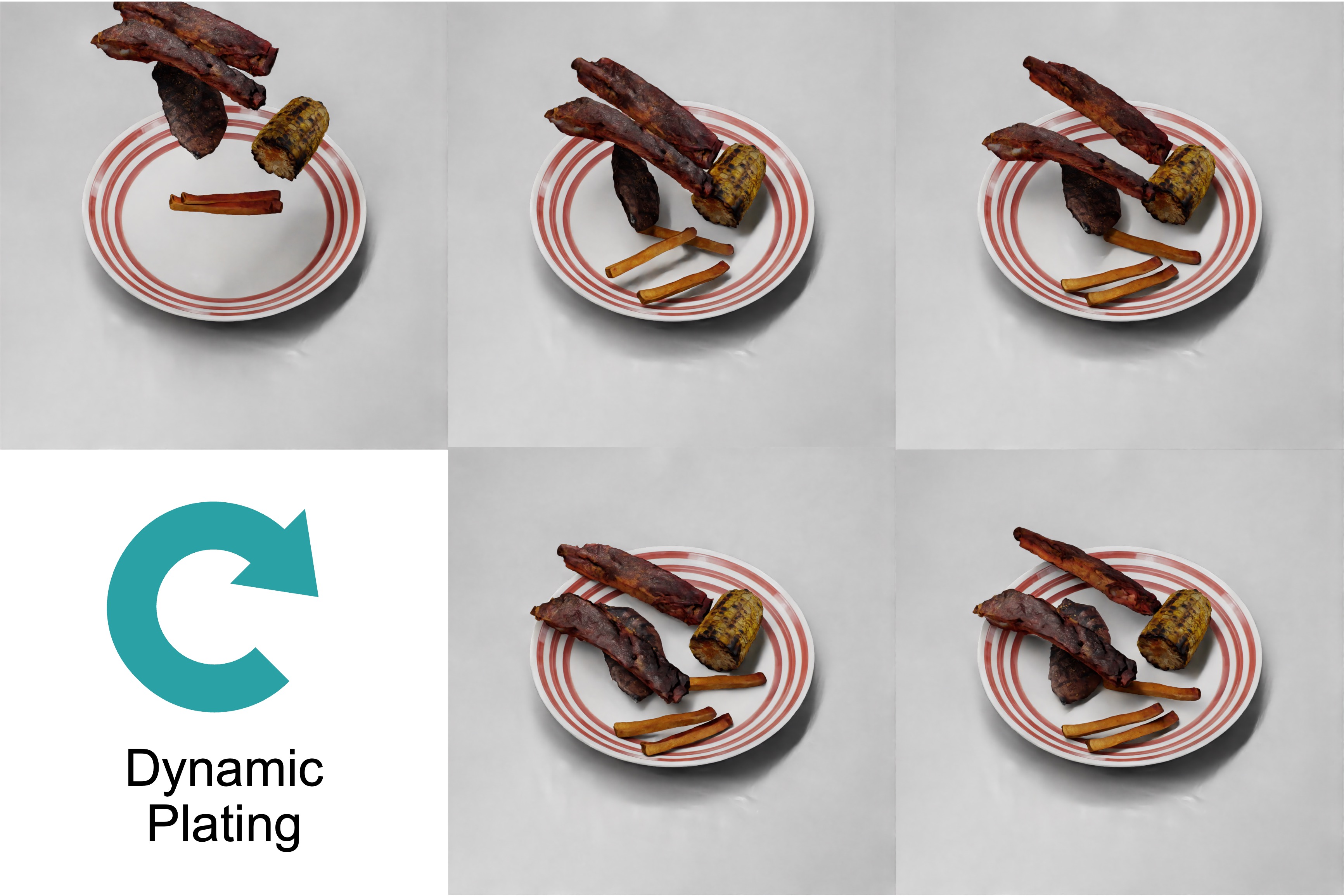}
    \end{minipage}%
    \begin{minipage}{.5\textwidth}
        \centering
        \includegraphics[width=.9\linewidth]{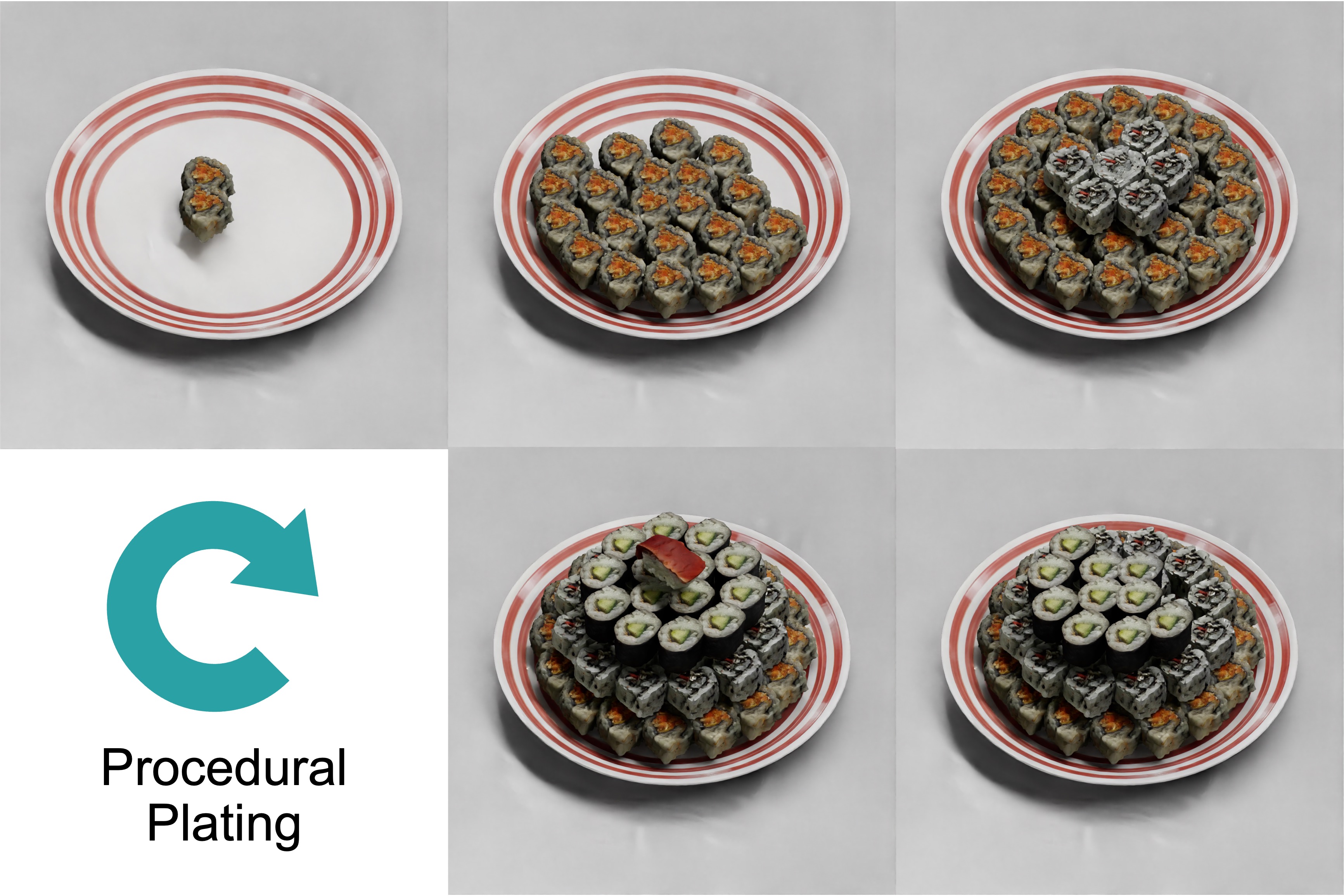}
    \end{minipage}
    \caption{(Left) Frames from a simulation showing randomly selected food items dropped onto a plate via dynamic plating. (Right) Frames from a procedural plating simulation, showcasing user-defined scene compositions and placement rules.}
    \label{fig:plating_methods}
\end{figure}

Our NV-Synth data generation framework leverages the Isaac Sim physics engine from NVIDIA Omniverse \cite{issac-sim} for GPU-accelerated simulation and uses high-resolution 3D food models from the NutritionVerse-3D dataset \cite{nutritionverse-3d}. To allow both scalable domain-randomized data generation and flexibly customizing scene compositions, we support two methods for generating synthetic meal images: dynamic plating and procedural plating. A visual comparison of the two methods can be seen in Figure~\ref{fig:plating_methods}.

\subsection{Dynamic Plating}

\begin{figure}[h]
    \centering
    \begin{minipage}{.5\textwidth}
        \centering
        \includegraphics[width=.9\linewidth]{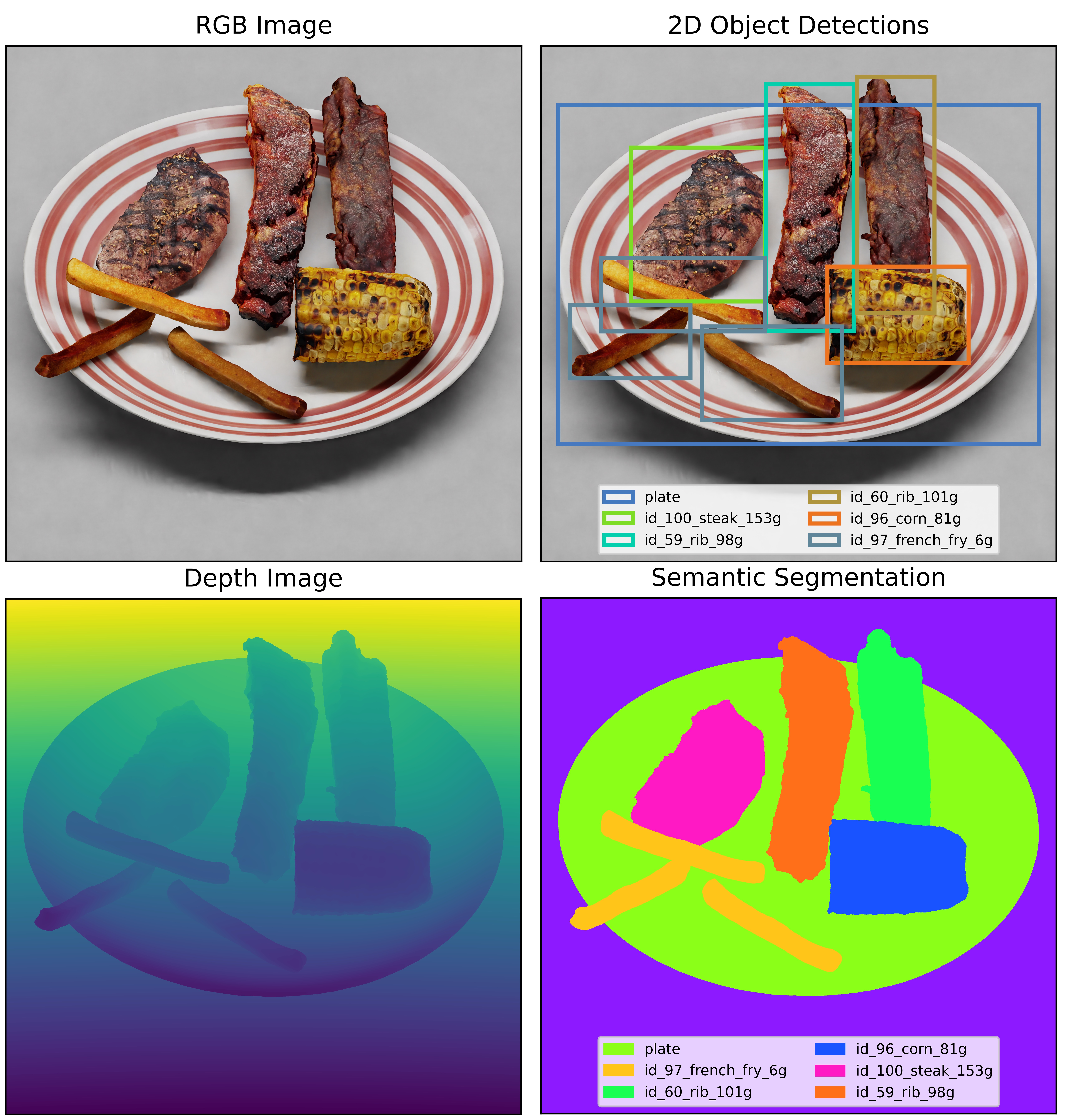}
    \end{minipage}%
    \begin{minipage}{.5\textwidth}
        \centering
        \includegraphics[width=.9\linewidth]{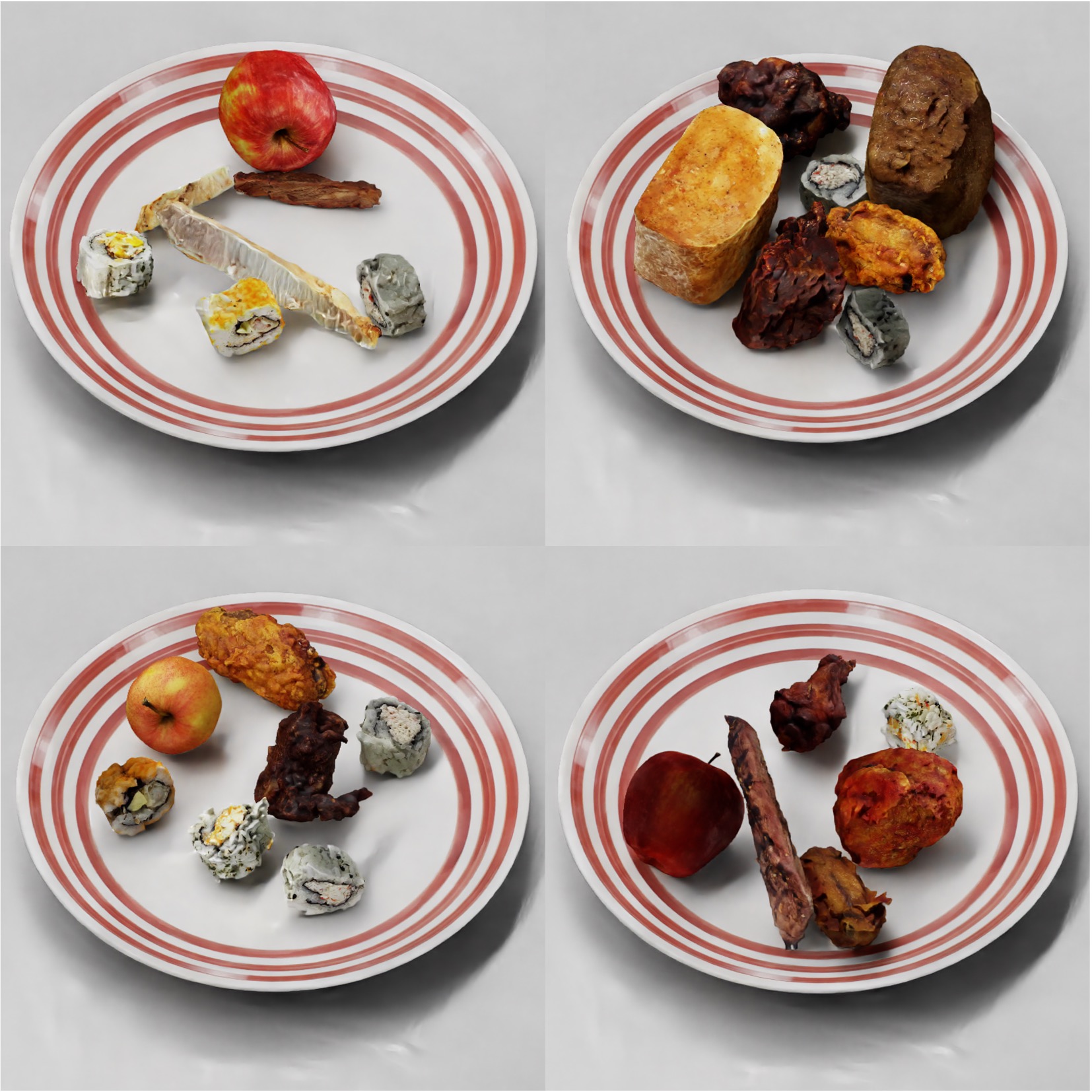}
    \end{minipage}
    \caption{(Left) A dynamically plated 3D food scene with multi-modal data and annotations, composed and rendered using NV-Synth. (Right) Randomly composed food scenes from NutritionVerse-3D dataset~\cite{nutritionverse-3d}, generated with dynamic plating.}
    \label{fig:dynamic_compositions}
\end{figure}

For dynamic plating, our framework generates scenes by randomly selecting up to seven different food items and simulating their placement onto a plate. These items are virtually dropped from varying heights, with random positions and orientations, allowing the physical properties of each food, such as mass and shape, to naturally influence their final arrangement. This process ensures variability in the meal presentations, enabling better domain randomization over the distribution of food scene compositions.

\subsection{Procedural Plating}
Alternatively, procedural plating allows for scene creation based on specific user-defined rules defined in a YAML configuration file. This mode gives users the ability to create meal scenes that follow particular dietary guidelines or plating styles, offering a more controlled and repeatable simulation environment. Unlike dynamic plating which treats each food item as a rigid body and allows interactions (like collisions) between food items in the scene, procedural plating fixes the items in the scene to a predefined pose parameterized by a user-configured rule. Figure~\ref{fig:procedural_plating_multi_modal} shows an example of two procedurally plated scenes from different views, along with their corresponding annotations. 

\subsection{Dataset Creation}
For creating the NV-Synth dataset, we use the dynamic plating method. We start by dividing the plate into segments to determine where food items should fall, reducing the chance of items overlapping or causing simulation errors due to being too close. If an item does not settle as expected or falls off the plate, it is removed, ensuring only well-placed items are included in our dataset. Figure~\ref{fig:dynamic_compositions} shows examples of four randomly composed scenes that were generated using dynamic plating as well as another dynamically plated scene and its corresponding multimodal annotations.

\begin{figure}[h]
    \begin{center}
        \includegraphics[width=0.6\linewidth]{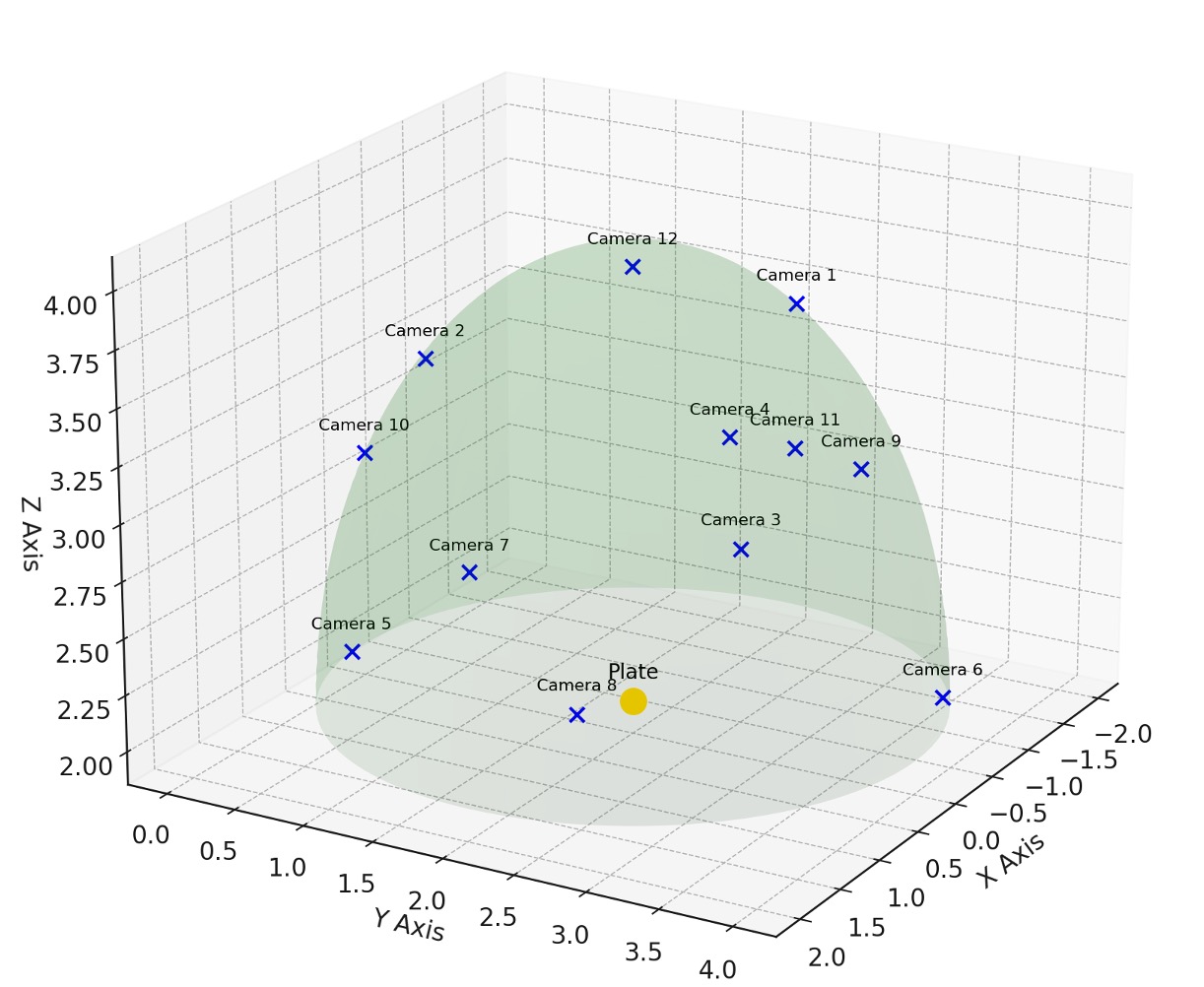}
    \end{center}
    \caption{Visualization of our camera placements with respect to the plate.}
    \label{fig:camera_placements}
\end{figure}

As shown in Figure~\ref{fig:camera_placements}, we record the scenes using cameras placed at 12 points on a hemisphere, based on a modified Fibonacci sphere to maximize distinct angles with low redundancy. From these, we select four random viewpoints for each scene to provide a varied perspective and to avoid bias in how the meals are viewed. We also adjust the lighting for each item by randomly varying the brightness of each item's texture between 1-2x. This helps add to the realism of the rendered images by mimicking different indoor lighting conditions and camera exposure settings. Additionally, focal lengths for the cameras are also varied in order to simulate different levels of zoom.

The result is the NV-Synth dataset, a collection of meal images rendered to look realistic, accompanied by multi-modal ground truth annotations for RGB, depth, 2D and 3D bounding boxes, and instance, amodal, and semantic segmentation masks. We compute nutritional metadata for each meal by aggregating nutritional information of the individual food items in the scene, based on data from the NutritionVerse-3D model library.

\section{Results and Discussion}
Using this pipeline, we create the NutritionVerse-Synth (NV-Synth) dataset, comprising 84,984 high-resolution 2D food images across 7,082 unique procedurally generated scenes. Each scene contains rich nutritional annotations, including mass, calories, carbohydrates, fats, and protein contents, and food item labels indicating the ingredients present in each dish. In addition, NVIDIA Omniverse \cite{issac-sim} also provides multi-modal ground truth annotations, including RGB, depth, bounding boxes, and segmentation masks. 

\begin{figure}[h]
    \begin{center}
        \includegraphics[width=0.8\linewidth]{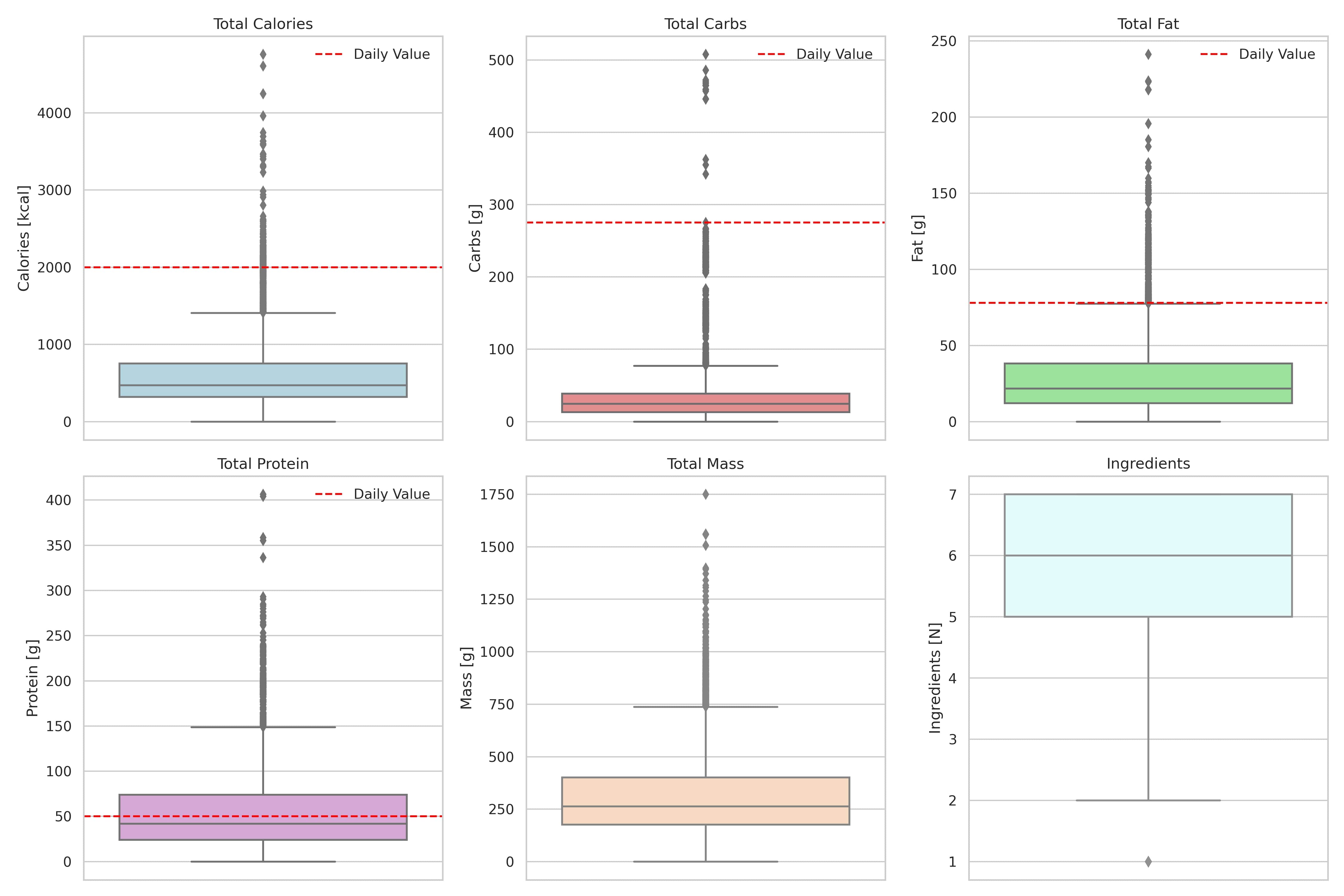}
    \end{center}
    \caption{Distribution of various nutritional factor amounts, mass, and ingredient counts in the generated NV-Synth food scenes. Recommended daily value is also shown in red for the nutrients.}
    \label{fig:nutritional_distribution}
\end{figure}

\begin{figure}[h]
    \begin{center}
        \includegraphics[width=0.8\linewidth]{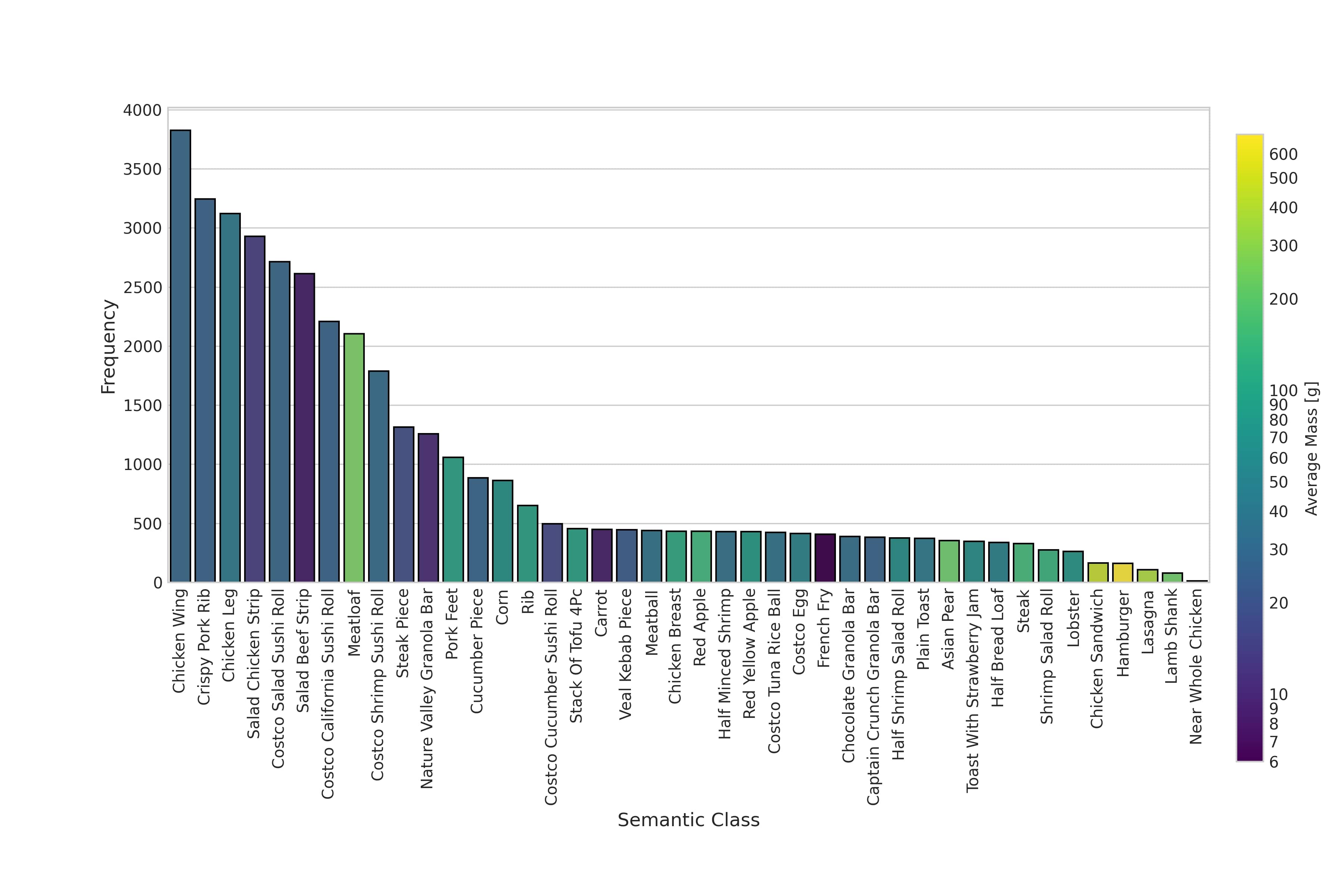}
    \end{center}
    \caption{Frequency distribution of the various semantic class types present across generated NV-Synth scenes along with the average mass of each food item type.}
    \label{fig:semantic_class_distribution}
\end{figure}

NV-Synth represents 105 distinct foods (45 unique semantic types as shown in Figure~\ref{fig:semantic_class_distribution}). Each food item appears in ~369 scenes on average, and every scene contains an average of 5.62 items. Figure~\ref{fig:nutritional_distribution} details the mean nutritional contents across scenes. We also release a 28,328 image subset for training/validation/testing (60\%/20\%/20\% split), created by randomly sampling 4 of 12 viewpoints per scene to reduce bias.

\section{Conclusion}

In this work, we have presented NutritionVerse-Synth (NV-Synth), a novel large-scale synthetic dataset to advance food image analysis and dietary assessment. NV-Synth represents the largest and most comprehensive synthetic food dataset to date. The scale, realism, and detailed annotations unlock new opportunities for developing and rigorously evaluating computer vision techniques for dietary assessment and food recognition. By releasing the dataset and simulation pipeline publicly, we hope to provide an essential resource to accelerate nutrition-focused research and applications. Future work includes continued expansion of the dataset and integration with other food image collections.

\vspace{-0.1in}
\begin{ack}
This work was supported by the National Research Council Canada (NRC) through the Aging in Place (AiP) Challenge Program, project number AiP-006. 
\end{ack}

{
\small

\bibliography{refs}

\begin{thebibliography}{1}

\bibitem{malnutrition-qol}
Heather~H. Keller, Truls Østbye, and Goy Richard.
\newblock Nutritional risk predicts quality of life in elderly community-living canadians.
\newblock {\em The Journals of Gerontology: Series A}, 59(1):M68–M74, 2004.

\bibitem{kipnis2003structure}
Victor Kipnis, Amy~F Subar, Douglas Midthune, Laurence~S Freedman, Rachel Ballard-Barbash, Richard~P Troiano, Sheila Bingham, Dale~A Schoeller, Arthur Schatzkin, and Raymond~J Carroll.
\newblock Structure of dietary measurement error: results of the open biomarker study.
\newblock {\em American journal of epidemiology}, 158(1):14--21, 2003.

\bibitem{10.1145/3347448.3357172}
Yoshikazu Ando, Takumi Ege, Jaehyeong Cho, and Keiji Yanai.
\newblock Depthcaloriecam: A mobile application for volume-based foodcalorie estimation using depth cameras.
\newblock In {\em Proceedings of the 5th International Workshop on Multimedia Assisted Dietary Management}, MADiMa '19, page 76–81, New York, NY, USA, 2019. Association for Computing Machinery.

\bibitem{thames2021nutrition5k}
Quin Thames, Arjun Karpur, Wade Norris, Fangting Xia, Liviu Panait, Tobias Weyand, and Jack Sim.
\newblock Nutrition5k: Towards automatic nutritional understanding of generic food.
\newblock In {\em Proceedings of the IEEE/CVF Conference on Computer Vision and Pattern Recognition}, pages 8903--8911, Nashville, 2021. IEEE.

\bibitem{nutritionverse-3d}
Chi-en~Amy Tai, Matthew Keller, Mattie Kerrigan, Yuhao Chen, Saeejith Nair, Pengcheng Xi, and Alexander Wong.
\newblock Nutritionverse-3d: A 3d food model dataset for nutritional intake estimation.
\newblock In {\em Conference on Computer Vision and Pattern Recognition (CVPR)}, Women in Computer Vision (WiCV), Vancouver, 2023. IEEE.

\bibitem{issac-sim}
NVIDIA.
\newblock Nvidia isaac sim, 2023.

\end{thebibliography}
}

\end{document}